\documentclass[runningheads]{llncs}
\usepackage{graphicx}

\usepackage{tikz}
\usepackage{comment}
\usepackage{amsmath,amssymb}

\usepackage{graphicx}
\usepackage{amsmath}
\usepackage{amssymb}
\usepackage{booktabs}
\usepackage{mathrsfs}
\usepackage{diagbox}
\usepackage{multirow}
\usepackage{makecell}
\usepackage{epsfig}
\usepackage{graphicx} 
\usepackage[breaklinks=true,bookmarks=false]{hyperref}
\hypersetup{
    colorlinks=true,
    citecolor=darkblue
}
\definecolor{darkblue}{rgb}{0,0.08,0.45}
\newcommand{\approach}{iCAR}

\usepackage{color, colortbl}
\newcommand{\demph}[1]{\textcolor[rgb]{0.565,0.565,0.565}{#1}}
\makeatletter
\renewcommand*{\@fnsymbol}[1]{\ensuremath{\ifcase#1\or \dagger \or * \or \ddagger\or
    \mathsection\or \mathparagraph\or \|\or **\or \dagger\dagger
    \or \ddagger\ddagger \else\@ctrerr\fi}}
\makeatother
\usepackage[accsupp]{axessibility}  % Improves PDF readability for those with disabilities.

\begin{document}
% \renewcommand\thelinenumber{\color[rgb]{0.2,0.5,0.8}\normalfont\sffamily\scriptsize\arabic{linenumber}\color[rgb]{0,0,0}}
% \renewcommand\makeLineNumber {\hss\thelinenumber\ \hspace{6mm} \rlap{\hskip\textwidth\ \hspace{6.5mm}\thelinenumber}}
% \linenumbers
\pagestyle{headings}
\mainmatter

\title{iCAR: Bridging Image Classification and Image-text Alignment for Visual Recognition}

%******************

\titlerunning{iCAR: Bridging Image Classification and Image-text Alignment ...}
% If the paper title is too long for the running head, you can set
% an abbreviated paper title here
%
\author{Yixuan Wei\inst{1}, Yue Cao\inst{2}\thanks{Contact person. Yixuan Wei, Zhuliang Yao and Zhenda Xie are long-term interns at MSRA.}, Zheng Zhang\inst{2}, Zhuliang Yao\inst{1},
\\Zhenda Xie\inst{1}, Han Hu\inst{2}, and Baining Guo\inst{2}}
\authorrunning{Y. Wei et al.}
% First names are abbreviated in the running head.
% If there are more than two authors, 'et al.' is used.
%
\institute{$^1$Tsinghua University~~~~
$^2$Microsoft Research Asia}%\\

%******************
\maketitle

%%%%%%%%% ABSTRACT
\begin{abstract}
   Image classification, which classifies images by pre-defined categories, has been the dominant approach to visual representation learning over the last decade. Visual learning through image-text alignment, however, has emerged to show promising performance, especially for zero-shot recognition. We believe that these two learning tasks are complementary, and suggest combining them for better visual learning. We propose a deep fusion method with three adaptations that effectively bridge two learning tasks, rather than shallow fusion through naïve multi-task learning. First, we modify the previous common practice in image classification, a linear classifier,  with a cosine classifier which shows comparable performance. Second, we convert the image classification problem from learning parametric category classifier weights to learning a text encoder as a meta network to generate category classifier weights. The learnt text encoder is shared between image classification and image-text alignment. Third, we enrich each class name with a description to avoid confusion between classes and make the classification method closer to the image-text alignment. We prove that this deep fusion approach performs better on a variety of visual recognition tasks and setups than the individual learning or shallow fusion approach, from zero-shot/few-shot image classification, such as the Kornblith 12-dataset benchmark, to downstream tasks of action recognition, semantic segmentation, and object detection in fine-tuning and open-vocabulary settings. The code will be available at \href{https://github.com/weiyx16/iCAR}{\color{magenta}https://github.com/weiyx16/iCAR}.
   
\end{abstract}

%%%%%%%%% BODY TEXT
\section{Introduction}
\label{sec:intro}

Image classification, a long-standing vision problem, has played as an important driving force for the remarkable success of deep learning in computer vision. The deep representations learnt through this task, such as using the ImageNet-1K dataset~\cite{deng2009imagenet} which involves 1,000 object categories, have been transferred to various vision tasks such as object detection, semantic segmentation, video classification, etc., to significantly advance relevant fields~\cite{girshick2014rich,long2015fully,tran2015learning}.

Recently, visual learning through image-text alignment has received more and more attention. The image-text alignment task treats an image and its associated alt-text as a positive pair and the image with all other alt-texts as negative ones. By contrasting positive and negative pairs, it learns visual representations and associates images with arbitrary semantics. This approach has been shown to seize the strong zero-shot classification capability~\cite{radford2021clip} and learn good visual representations~\cite{jia2021align}.

\begin{figure}[t]
    \centering
    \includegraphics[width=\linewidth]{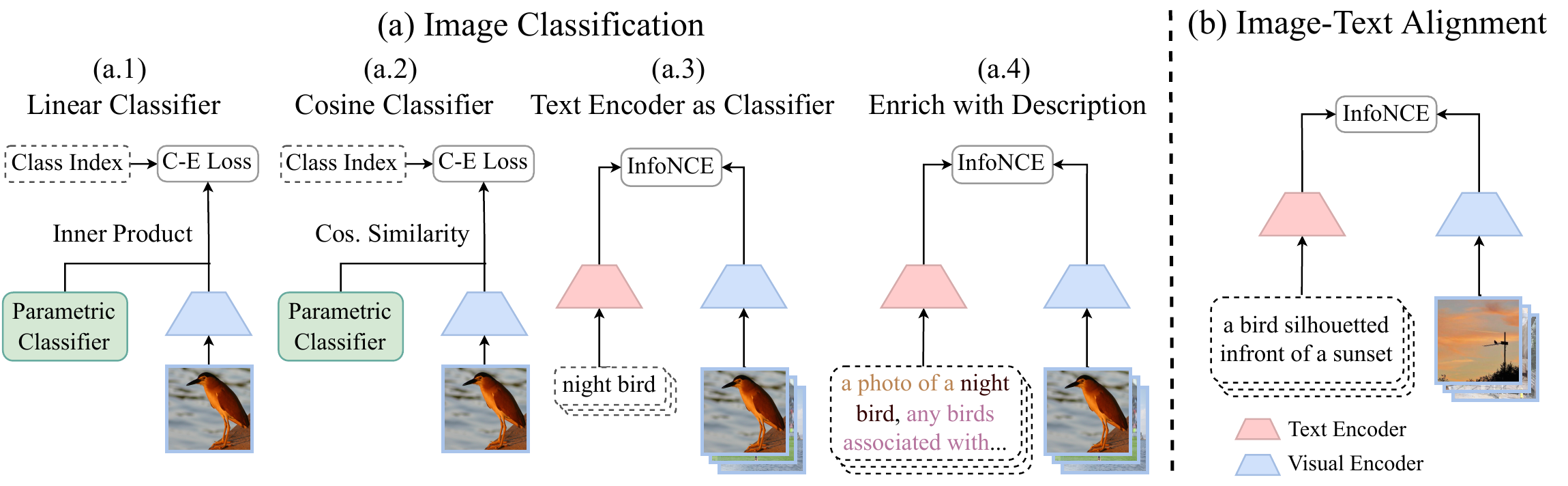}
    \caption{An illustration of our proposed approach to bring image classification task (see a) to image-text alignment task (see b). From the perspective of training loss, classifier type, and label/input granularity, we modify the linear classifier to a cosine classifier (see a.2), introduce a text encoder as a meta classifier (see a.3) and enrich each ambiguous category name with a meaningful description (see a.4). After the re-formulation, we are able to deeply unify two tasks and benefit from both.} 
    \label{fig:arch}
\end{figure}
In our view, these two learning methods have their own strengths and are essentially complementary. For example, annotations in image classification tasks are often precise, compact and consistent, and embody good properties for visual representation learning. However, there are also drawbacks, such as relatively small public available datasets and insufficient concepts coverage that categories are often limited to a pre-defined set, inaccurate or ambiguous class names (14.6\% of classes in ImageNet-22K have repetitions in class names, but each class represents a completely different visual concept, shown in Figure~\ref{fig:classname-repeat}).
These shortcomings of image classification in visual learning can be well complemented by the image-text alignment task, which can have good concepts coverage, semantically-rich sentences, and large-scale datasets with lower collection costs such as the Laion dataset including 400 million image-text pairs~\cite{schuhmann2021laion400m}. Conversely, the disadvantage of the image-text alignment task is that each image is described by a noisy and non-compact sentence. And it can be supplemented by a clean and compact image classification task.

While we note a straightforward solution is to combine both tasks within a naïve multi-task learning framework, we investigate a deep fusion approach that effectively unifies the two tasks from the perspective of training loss, classifier type, and label/input granularity.
First, we modify the previous common practice of using a linear classifier in image classification by a cosine classifier that shows competitive performance. Second, we convert the image classification problem from learning parametric classifier weights to learning a text encoder that acts as a meta network to generate category classifier weights. The image classification task and the image-text alignment task now perform in the same embedding space, with cosine distance as the metric, and the features are extracted from shared visual/text encoders. Third, we enrich each class name with a description to avoid misconceptions between classes and bring the classification method closer to the image-text alignment approach regarding the input granularity.

For example, after enrichment, ``a photo of a night bird, any bird associated with night: owl, nightingale, nighthawk'' has a similar granularity and expression, compared to a caption/sentence, and also provides more detailed information than the given class name ``night bird''. Figure~\ref{fig:arch} is an illustration on the adaptations of the image classification task to the image-text alignment task, so that they can be deeply fused. 

This deep fusion framework that combines image classification and image-text alignment allows for a wide range of applications, from straight-forward tasks such as image classification and image-text retrieval that align with the learning objectives in zero-shot/few-shot settings, to transferring to various vision tasks such as video action recognition, semantic segmentation, and object detection through fine-tuning. In particular, the fine-tuning is conducted not only on the vision network as before, but also on the text encoder, which empirically performs better, especially for down-stream tasks that require adjusting the text space to distinguish fine-grained categories. 

For evaluation of the straight-forward tasks of image classification and image-text retrieval, we first experiment with Conceptual Captions~\cite{sharma2018CC3m} and ImageNet-1K~\cite{deng2009imagenet}. 

Our deep fusion approach performs better than the individual learning or shallow fusion approach, on both the zero-shot and few-shot settings of the Kornblith 12-dataset benchmark~\cite{imagnettransfer}. This indicates that the deep fusion approach can better benefit from the strengths of both tasks. When using the training datasets of Laion-400M~\cite{schuhmann2021laion400m} and ImageNet-22K~\cite{deng2009imagenet}, \approach{} can surpass the previous state-of-the-art approaches in both zero-shot and few-shot settings for the standard 12-dataset benchmark, including CLIP~\cite{radford2021clip} and ALIGN~\cite{jia2021align}. 

With fine-tuning, we prove that the proposed approach has strong representation learning and open-vocabulary recognition capabilities when transferred to other vision tasks. The approach achieves 52.5 mIoU on ADE-20K validation set, by using a MaskFormer~\cite{cheng2021maskformer} framework and the Swin-B backbone, 0.4/0.6 higher than previous methods based solely on image-classification or image-text. What's more, there are huge improvements in open-vocabulary semantic segmentation performance compared to using the pre-trained models directly, such as 47.7 mIoU v.s. 21.3 mIoU on PASCAL-VOC~\cite{PASCAL}, 53.8 mIoU v.s. 10.0 mIoU on Cityscapes~\cite{cordts2016cityscapes} and 14.7 mIoU v.s. 2.1 mIoU on COCO Stuff~\cite{lin2014coco}. 

\section{Related Work}
\label{sec:related-work}

\paragraph{Visual recognition with multi-way discrete classifier}
Visual recognition is widely used in numerous vision tasks with different recognition granularity. Previous dominated methods all adopt the multi-way parametric classifier with the softmax cross-entropy loss, such as image-level classification~\cite{krizhevsky2012alexnet,szegedy2015googlenet,simonyan2014vgg,he2015resnet,xie2017resnext,dosovitskiy2020vit,Swin}, object-level classification in object detection~\cite{girshick2014rich,ren2015faster,he2017mask,carion2020detr}, pixel-level classification in semantic/instance segmentation~\cite{long2015fully,chen2017rethinking,yin2020DNL}, video-level action classification~\cite{tran2015learning,carreira2017i3d,qiu2017P3D,feichtenhofer2018slowfast,liu2021videoswin} and etc.
Some recent approaches~\cite{wu2018improving,khosla2020supervised} try to explore the non-parametric classifier for supervised classification, which performs the contrastive learning to attract the images within the same class and repel those belonging to different classes inside each batch.
However, all these approaches aren't aware of the semantic meaning of each class, thus they can not classify images belonging to a newly-giving set of classes, if these classes do not have a perfect match to the training classes.

\vspace{-.5em}
\paragraph{Visual-linguistic alignment} Leveraging natural language as supervision from image-text pairs for joint visual-linguistic representation learning is a long studied research topic~\cite{nips13devise,norouzi2013zero,joulin2016first-flickr-vl,gomez2017self,desai2021virtex,sariyildiz2020icmlm,zhang2020vl-medical,radford2021clip,jia2021align}. DeViSE~\cite{nips13devise} firstly introduces label semantic embedding to refactor image embedding after context-free multi-way classification. ICMLM~\cite{sariyildiz2020icmlm} and VirTex~\cite{desai2021virtex} drive the representation learning by predicting masked words in a sentence from corresponding image embedding in an autoregressive way. And ConVIRT ~\cite{zhang2020vl-medical} conducts visual-linguistic contrastive learning in medical-related datasets for task-specific visual representations. Due to the increasing of computation during these years, the used datasets are expanded from the small-scale datasets (e.g., Flickr~\cite{young2014flickr}, COCO Captions~\cite{lin2014coco}, Conceptual Captions 3M~\cite{changpinyo2021cc12m}) to web-scale ones (e.g., CLIP~\cite{radford2021clip}, ALIGN~\cite{jia2021align}, Laion~\cite{schuhmann2021laion400m}).
There are two pioneer works, CLIP~\cite{radford2021clip} and ALIGN~\cite{jia2021align}, which leverage the web-scale datasets with noisy image-alt-text pairs from the Internet and thus cover concepts with unlimited number during the training of image-text matching.
However, the crawled image-alt-text dataset is noisy, somewhat downgrading the visual recognition performance.

To the best of our knowledge, our paper is the first trial to deeply bridge the supervised classification and image-text alignment tasks, trying to design a new unified learning framework to benefit both. And we demonstrate that, with careful designs, image classification and image-text alignment could complement each other, and the proposed \approach{} could significantly outperform the single-task baselines and the shallow fusion approach with separate task heads.

In addition, there are still few study on how to leverage these pre-trained models for open-vocabulary recognition in downstream tasks.
In this paper, we present an extremely simple but effective baseline, to both load pre-trained visual encoder as backbone and the text encoder as visual classifier during fine-tuning on downstream tasks. This technique incurs competitive results on fine-tuned datasets, and remarkably good performance on open-vocabulary recognition for unseen datasets of the fine-tuned task.

\section{Approach}
\label{sec:approach}

\subsection{Preliminaries}

\noindent \textbf{Image classification}
Given a set of $<$image, category label index$>$ pairs, $\mathcal{D}=<I_i, C_i>_{i=1}^{|\mathcal{D}|}$, image classification task targets to predict the category label of a given image, usually through a visual encoder $f_v$, and a parametric category classifier $h_c$. The visual encoder transforms each raw image $I_i$ to an embedding $v_i$ in the feature space, and the classifier predicts the logit distribution $p_i$ over all pre-defined $N$ categories in $\mathcal{D}$, i.e. 1000 categories in ImageNet-1K dataset, from the embedding $v_i$. In most cases, the parametric category classifier $h_c$ is a weight matrix $W\in\mathcal{R}^{N\times H}$, where $H$ is the dimension of $v_i$ (for simplicity, bias term of $h_c$ is omitted). The logits $p_i$ of all categories are the inner product between $W$ and $v_i$, i.e. $p_i=W\cdot v_i$. Consider a given image $I_i$, a cross-entropy loss is applied between $p_i$ and $C_i$, and a complete formulation can be defined as:
\begin{equation}\label{eq:supervise}
    \mathcal{L}_{i} = -\log \frac{{\exp \left( { W_{C_i}\cdot f_v\left(I_i\right) } \right)}}{{\sum_{j=1}^{N} {\exp \left( {W_j\cdot f_v\left(I_i\right)} \right)} }},
\end{equation}
where $W_j$ is the parametric weight of $j$-th category. 

\noindent \textbf{Image-text alignment} 
Given a set of $<$image, caption$>$ pairs, $\mathcal{D}=<I_i, T_i>_{i=1}^{|\mathcal{D}|}$, image-text alignment task targets to close the distance of paired image-text but enlarge that of unpaired ones, through a visual encoder $f_v$ and a text encoder $f_t$. The visual encoder and text encoder transforms the image $I_i$ and the caption $T_i$ into embeddings $v_i$ and $s_i$, respectively. InfoNCE~\cite{InfoNCE}, a contrastive loss function is often applied to shrink the cosine distance of $v_i$ and $s_i$. Consider an image embedding $v_i$, a formulation of contrastive learning loss is defined as:
\begin{equation}\label{eq:vl-contrastive}
    \mathcal{L}_{i} = -\log \frac{{\exp \left( \cos\left( f_t\left(T_i\right), v_i \right)/\tau \right)}}{{\sum\limits_{{T_j} \in T} {\exp \left( \cos\left( f_t\left(T_j\right), v_i \right)/\tau \right)} }},
\end{equation}
where $\cos(\cdot, \cdot)$ denotes the cosine similarity between two vectors, $T$ is all the captions in batch, including one positive paired caption and $|T|-1$ negative ones, and $\tau$ is the temperature hyper-parameter to scale the logits.

\subsection{Bridging Image classification and Image-text alignment}
\label{sec:bridge}
To bridge image classification task and image-text alignment task for better unification, we propose three adaptations to align the training loss, unify the classifier and minimize the label granularity gap.

\noindent \textbf{Cosine classifier} 
As the formulation in Eqn.~\ref{eq:supervise}, the original image classification loss is a cross-entropy loss on top of the inner product similarity between the embedding $v_i$ and the parametric classifier $h_c$. This formulation isn't in line with the InfoNCE loss, shown in Eqn.~\ref{eq:vl-contrastive}, which is prevalent in image-text alignment task. We review the image classification task from the perspective of metric learning, and formulate it using a cosine classifier. To be more specific, we apply L2 normalization both on the parametric category classifier $h_c$ and the embedding $v_i$, and the optimization target is switched to maximize the cosine similarity of image features and their corresponding class features. We also scale the logits using a temperature $\tau$ to be consistent with InfoNCE and the cosine classifier based image classification loss is like this: 

\begin{equation}\label{eq:supervise-cosine}
    \mathcal{L}_{i} = -\log \frac{{\exp \left( \cos\left( W_{C_i}, v_i \right)/\tau \right)}}{{\sum_{j=1}^{N} {\exp \left( \cos\left( W_{j}, v_i \right)/\tau \right)} }},
\end{equation}

\begin{figure}[t]
\begin{minipage}[c]{0.55\textwidth}
    \centering
    \includegraphics[width=0.8\linewidth]{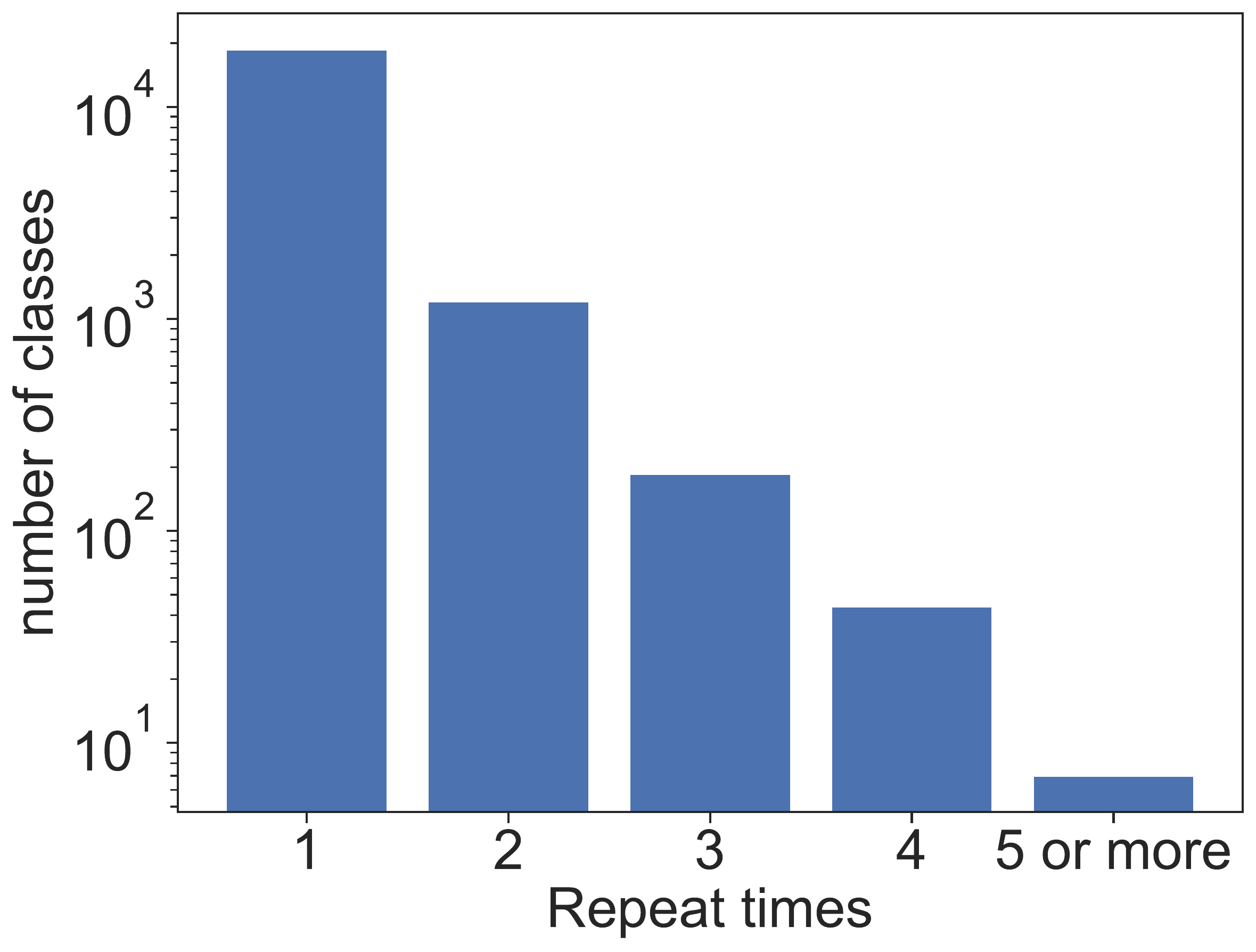}
\end{minipage}
\begin{minipage}[c]{0.4\textwidth}
    \caption{Statistical analysis on label names of 21843 categories in ImageNet-22K dataset. Only 18657 categories use unique label names, and thousands of categories have the same label names as at least one another category.}
    \label{fig:classname-repeat}
\end{minipage}
\end{figure}
We prove that a cosine classifier based approach can reach an on par performance with the traditional classification method (see Table~\ref{tab-ablation-cosine}).

\noindent \textbf{Text encoder as a meta network for image classification} 
Re-formulating the classification loss using a cosine classifier aligns the loss formats. However, the label information from two tasks, categories and captions respectively, are not shared between the parametric category classifier $h_c$ and the text encoder $f_t$. As shown in Section~\ref{sec:ablation}, combining two tasks shallowly leads to sub-optimal results, without benefiting much from the accurate annotation in image classification and rich concepts and open-vocabulary ability in image-text alignment. 

To this end, we take label semantics into consideration and propose to utilize the text encoder $f_t$ as a meta classifier for image classification. Formally, given a pair of $<$image $I_i$, category label index $C_i>$, we replace the $C_i$ with its corresponding class name $M_i$, for example, \textit{tench} for the 1st category in ImageNet-1K dataset. Furthermore, we adopt a text encoder $f_t$ on $M_i$ to generate the classifier weight on-the-fly, instead of optimizing a parametric category classifier $h_c$. The new formulation is shown as:
\begin{equation}\label{eq:supervise-text}
    \mathcal{L}_{i} = -\log \frac{{\exp \left( \cos\left( f_t\left(M_i\right), v_i \right)/\tau \right)}}{{\sum_{j=1}^{N} {\exp \left( \cos\left( f_t\left(M_j\right), v_i \right)/\tau \right)} }}.
\end{equation}
Note that, integrating the text encoder into image classification enables open-vocabulary ability and shrinks the gap between two tasks.

\begin{figure}[t]
\begin{minipage}[c]{0.55\textwidth}
    \centering
    \includegraphics[width=0.9\linewidth]{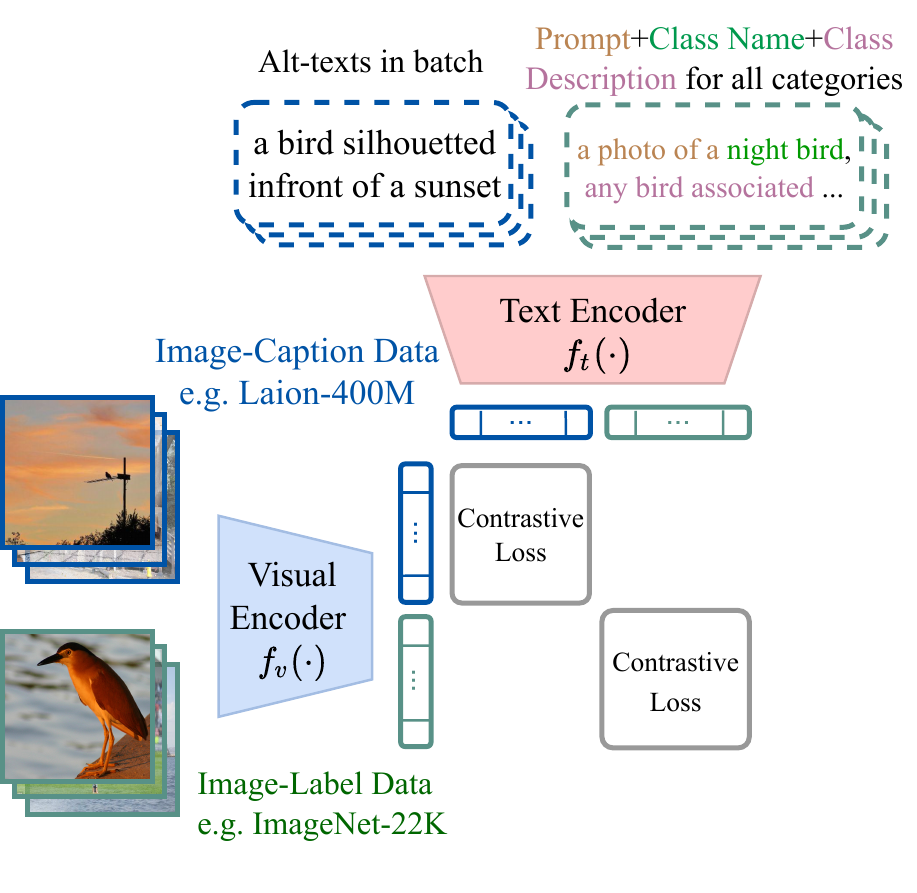}
\end{minipage}
\begin{minipage}[c]{0.4\textwidth}
    \caption{The detailed illustration of our unified training method. \approach{} performs both the image classification (in green) and image-text alignment (in blue) in a unified contrastive learning scheme. \emph{Best viewed in color}.}
    \label{fig:arch-details}
\end{minipage}
\end{figure}

\noindent \textbf{Enriched class name with description} 
Replacing label index $C_i$ with label name $M_i$ has largely bridged the image classification and image-text alignment. To further minimize the label granularity gap between label names (one or two words) and image captions (a complete sentence), we propose to integrate the detailed description $D_i$ of each category. The description can be found from corresponding synset in wordnet~\cite{miller-1994-wordnet} for ImageNet dataset or the beginning sentence of the corresponding Wikipedia page. We also add a prompt to make the sentence more fluent. So, as shown in Figure~\ref{fig:arch-details}, the label for each category is formed through the following template: \textit{prompt sentence+category name+description}, and due to simplicity and similarity, we annotate it as $T_i$. The formulation of our proposed image classification framework is shown as
\begin{equation}\label{eq:supervise-text-des}
    \mathcal{L}_{i} = -\log \frac{{\exp \left( \cos\left( f_t\left(T_i\right), v_i \right)/\tau \right)}}{{\sum_{j=1}^{N} {\exp \left( \cos\left( f_t\left(T_j\right), v_i \right)/\tau \right)} }}.
\end{equation}

The detailed description enables a deeper understanding of each category and reduces the misconception error, especially when only the class name is known without further details. For example, in ImageNet-22K dataset, at least 6 categories are labeled as \emph{jack} but representing 6 various visual concepts, e.g., one means any of several fast-swimming predacious fishes of tropical to warm temperate seas, like Almaco jack, and 
another one means a tool for exerting pressure or lifting. As shown in Figure~\ref{fig:classname-repeat}, over 1,000 class names in ImageNet-22K dataset repeat twice and hundreds repeat three times or more.
Also, the corresponding description with its class name would have similar granularity to captions, and thus bring the classification method closer to the image-text alignment.

\subsection{A unified framework}

We propose three methods to bridge image classification task and image-text alignment task, from the perspective of training loss, classifier type and label/input granularity. The image classification is re-formulated in Eqn.~\ref{eq:supervise-text-des}, and it is finally in line with the InfoNCE loss (see Eqn.~\ref{eq:vl-contrastive}) in image-text alignment. 

Here, we present the unified contrastive learning loss, which is added on top of the visual and text encoders, to simultaneously perform the tasks of image classification and image-text alignment. A detailed illustration of our unified contrastive learning loss is in Figure~\ref{fig:arch-details}. A general formulation is defined as:
\begin{equation}\label{eq:uni-contrastive}
\small
    L\left( \mathcal{D} \right) = -\frac{1}{|\mathcal{D}|}\sum\limits_{({I_i},{T_i}) \in \mathcal{D}} {\log \frac{{\exp \left({\cos\left( {f_t}\left( {{T_i}} \right), {f_v}\left( {{I_i}} \right) \right)/\tau } \right)}}{{\sum\limits_{{T_j} \in T} {\exp \left( {\cos\left( {f_t}\left( {{T_j}} \right), {f_v}\left( {{I_i}} \right)  \right)/\tau } \right)} }}},
\end{equation}
where $\mathcal{D}$ is the combination of image-text alignment and image classification datasets. Notice that, as mentioned in section~\ref{sec:bridge}, we discard the original category label index $C_i$ in image classification dataset and adopt a complete sentence $T_i$, including \textit{prompt sentence+category name+description}, as the annotation for each category. $T$ in Eqn.~\ref{eq:uni-contrastive} is the whole text set for contrastive learning (e.g., all captions in batch or all class labels with descriptions), $f_v(\cdot)$ and $f_t(\cdot)$ denotes the visual and text encoders and $\cos(\cdot,\cdot)$ denotes cosine similarity metric. $\tau$ is the temperature hyper-parameter to scale the logits as in InfoNCE.

\subsection{Applications}
\begin{figure}[t]
    \centering
    \includegraphics[width=\linewidth]{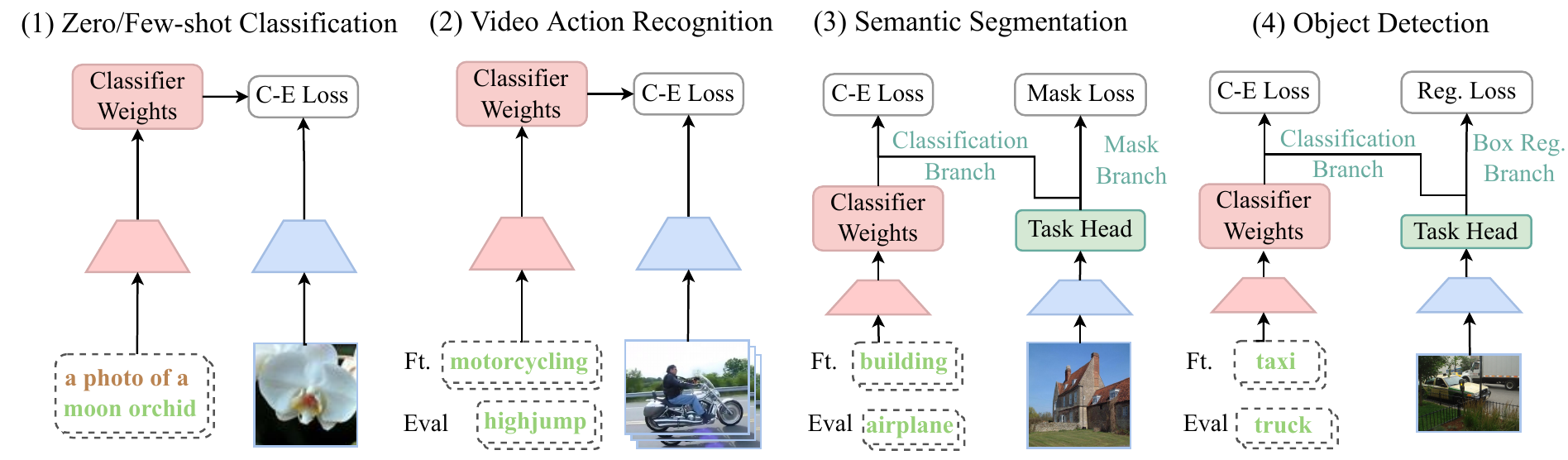}
    \caption{In our evaluation, we also integrate the pre-trained text encoder as a meta visual classifier in every scenarios. We find it serves as a good initialization and helps few-shot learning, and enables open-vocabulary recognition in downstream tasks.}
    \label{fig:arch-application}
\end{figure}
After the unified training, we evaluate our model on numerous applications, such as \emph{zero-shot evaluation}, \emph{few-shot learning}, and \emph{fine-tuning on downstream tasks} (e.g., semantic segmentation, action recognition, and object detection).
As shown in Figure~\ref{fig:arch-application}, for applications, our major philosophy is to load not only the visual encoder as backbone (as blue) but also the text encoder as the meta visual classifier (as red) from the pre-trained model. 

For few-shot learning, as the number of training samples for each class is very small (e.g., 1 or 4 samples per class), the text encoder could serve as a good initial classifier. This would alleviate the situation in~\cite{radford2021clip} that the performance of the few-shot setting is even worse than that of the zero-shot setting.

For other downstream tasks, previous approaches adopt the task specific framework with only the pre-trained visual encoder as initialization, and use a discrete multi-way classifier on top for recognition. Here we replace the original classifier with the pre-trained text encoder to generate the weight of each class with class names as input.
After fine-tuning, we find that the models have surprisingly good open-vocabulary recognition capability on unseen datasets of the same fine-tuning task (see Section~\ref{exp:finetune}). The reason may be that the text encoder still maintains part of open-vocabulary capability learned in the pre-training.

\section{Experiments}
\label{sec:exp}
\subsection{Ablation Study}\label{sec:ablation}
\noindent \textbf{Setup}
We use Conceptual Cations 3M (CC)~\cite{sharma2018CC3m} as image-text alignment dataset and ImageNet-1K (IN-1K)~\cite{deng2009imagenet} as image classification dataset. In all our ablation studies, we adopt a Swin-tiny~\cite{Swin} as our visual encoder and a RoBERTa-base~\cite{liu2019roberta} as our text encoder.
We train the models for 100 epochs on IN-1K (or equivalently on CC) with a batch size of 128 per GPU on 8 V100 GPUs in total. For approaches utilizing both datasets, we sample half number of images from both in each mini-batch. An AdamW optimizer~\cite{kingma2014adam} with a base learning rate of 2e-4, weight decay of 0.01, RandAugment~\cite{cubuk2020randaugment} and stochastic depth~\cite{huang2016deep} of 0.1 are used. The learning rate schedule is composed of a linear warm-up for 5 epochs and a cosine learning rate decay.
For direct/zero-shot classification, we adopt several variants of IN-1K validation sets as the benchmark for comprehensive understanding, such as IN, IN-V2~\cite{imagenetv2}, IN-Rendition (IN-R)~\cite{hendrycks2021many}, IN-Sketch (IN-S)~\cite{wang2019learningSketch}, and also the widely-used Kornblith 12-dataset benchmark.
We follow CLIP~\cite{radford2021clip} on the test set splits and evaluation metrics, and test the zero-shot classification ability of the pre-trained model. We borrow the same prompt list as CLIP and ensemble the results for fair comparison. 
For few-shot classification, we follow~\cite{radford2021clip} to adopt Kornblith 12-dataset benchmark.
For zero-shot cross-modal retrieval, we evaluate the models on Flickr and MSCOCO with the standard settings in~\cite{radford2021clip}.

\noindent \textbf{Ablation: cosine classifier} 
Firstly, we compare the cosine classifier in Eqn.~\ref{eq:supervise-cosine} (with temperature $\tau$ as 0.05) to original linear classifier in Eqn.~\ref{eq:supervise} on IN-1K image classification task. 

In Table~\ref{tab-ablation-cosine}, we observe that the cosine classifier performs competitive to an inner product based classifier on all validation sets, and thus it supports us to bridge image classification task to image-text alignment task using a cosine classifier by default.

\begin{table}[t]
  \centering
  \addtolength{\tabcolsep}{+1.8pt}
  \begin{tabular}{c|cccc}
    \toprule
    \multicolumn{1}{c}{} & \multicolumn{4}{c}{ImageNet-Related}  \\
    \cmidrule(lr){2-5}
    Method & IN & IN-V2 & IN-R & IN-S \\
    \hline
    Linear Classifier & 80.9 & 69.5 & 42.9 & 29.4  \\ 
    Cosine Classifier  & 81.5 & 69.9 & 43.5 & 31.1 \\
    \bottomrule
  \end{tabular}
  \caption{Ablation study with the cosine classifier in image classification task on ImageNet related benchmarks (top-1 accuracy).}
  \label{tab-ablation-cosine}
\end{table}

\noindent \textbf{Ablation: text encoder as a meta classifier} Here we ablate whether to use text encoder as a meta classifier for image classification, in a joint learning setting which both performs image classification on IN-1K and image-text alignment on CC. Our results are in Table~\ref{tab-ablation-text-enc}. Both \emph{Split head} and \emph{Text encoder} approaches adopt Eqn.~\eqref{eq:vl-contrastive} for image-text alignment. But \emph{Split head} adopts Eqn.~\eqref{eq:supervise} for image classification, which means that it performs two tasks in separate heads. \emph{Text encoder} adopts Eqn.~\eqref{eq:supervise-text} for image classification, which utilizes the text encoder as a meta classifier and shares the text encoder among two tasks.  

As shown in Table~\ref{tab-ablation-text-enc}, we observe that adopting text encoder could benefit from both tasks to achieve better performance on zero-shot classification of IN-R and IN-S, and zero-/few-shot classification on the 12-dataset benchmarks.

\begin{table}[t]
  \centering
  \small
  \addtolength{\tabcolsep}{+1.8pt}
  \renewcommand{\arraystretch}{0.95}
  \begin{tabular}{c|cccc|cccc}
    \toprule
    \multicolumn{1}{c}{} & \multicolumn{4}{c}{ImageNet-Related} & 
    \multicolumn{4}{c}{12-dataset Avg.} \\
    \cmidrule(lr){2-5}\cmidrule(lr){6-9}
    Method & IN & IN-V2 & IN-R & IN-S & 0 Shot & 1 Shot & 4 Shot & 16 Shot \\
    \hline
    Split head-Sup & 80.6 & 69.1 & 49.2 & 38.3 & - & - & - & - \\
    Split head-Text & 45.0 & 38.5 & 41.8 & 24.7 & 35.1 & 42.7 & 54.9 & 65.4 \\
    Text encoder (w/o Desc.) & 80.5 & 69.1 & 49.8 & 38.6 & 37.7 & 44.2 & 55.8 & 66.1 \\
    Text encoder (w. Desc.) & 80.4 & 69.1 & 49.7 & 38.7 & 39.1 & 46.3 & 57.0 & 66.9 \\
    \bottomrule
  \end{tabular}
  \caption{Ablation study on whether adopting text encoder and integrating description for image classification on ImageNet series and zero-/few-shot evaluation on 12-dataset benchmark (top-1 accuracy).}
  \label{tab-ablation-text-enc}
\end{table}

\noindent \textbf{Ablation: enriched category name with description}
Here we ablate whether to enrich class names with descriptions in a joint learning setting, using Eqn.~\eqref{eq:supervise-text} (w/o Desc.) and Eqn.~\eqref{eq:supervise-text-des} (w. Desc.) for image classification task, respectively. Besides evaluating on zero-/few-shot classification, we also perform this ablation study on the cross-modal retrieval tasks of Flickr and MSCOCO.

We observe that enriching each class name with its description could reduce the misalignment in the class names to benefit the classification capability (1.4+ on averaged accuracy of 12-dataset benchmark in Table~\ref{tab-ablation-text-enc}), and bridge the input gap of two tasks to benefit the cross-modal retrieval or image-text alignment capability (see Table~\ref{tab-ablation-enrich-retrieval}).

\begin{table}[t]
  \centering
  \small
\addtolength{\tabcolsep}{0.9pt}
\renewcommand{\arraystretch}{0.85}
  \begin{tabular}{ccccccccccccc}
    \toprule
     & \multicolumn{6}{c}{Text Retrieval} & \multicolumn{6}{c}{Image Retrieval} \\
    &  \multicolumn{3}{c}{Flickr} &  \multicolumn{3}{c}{MSCOCO}&  \multicolumn{3}{c}{Flickr} &  \multicolumn{3}{c}{MSCOCO} \\
    \cmidrule(lr){2-4} \cmidrule(lr){5-7} \cmidrule(lr){8-10} \cmidrule(lr){11-13}
    & R@1 & R@5 & R@10 &R@1 & R@5 & R@10 &R@1 & R@5 & R@10 &R@1 & R@5 & R@10  \\
    \midrule
    w/o Desc. &48.7&79.9&88.5& 25.3&50.5&62.9& 40.5&70.3&80.7&  17.9&39.8&51.7 \\
    w Desc. &51.4&81.7&89.0& 28.2&53.0&65.7& 41.0&71.2&81.1&  20.4&44.2&56.4 \\
    \bottomrule
  \end{tabular}
  \caption{Ablation study on whether enriching category names with descriptions of zero-shot cross-modal retrieval performance on MSCOCO and Flickr.}
  \label{tab-ablation-enrich-retrieval}
\end{table}

\begin{table}[t]
  \centering
  \small
  \addtolength{\tabcolsep}{+1.8pt}
  \renewcommand{\arraystretch}{0.95}
  \begin{tabular}{c|cccc|cccc}
    \toprule
    \multicolumn{1}{c}{} & \multicolumn{4}{c}{ImageNet-Related} & 
    \multicolumn{4}{c}{12-dataset Avg.} \\
    \cmidrule(lr){2-5}\cmidrule(lr){6-9}
    Method & IN & IN-V2 & IN-R & IN-S & 0 Shot & 1 Shot & 4 Shot & 16 Shot \\
    \hline
    Sup-only  & 80.9 & 69.5 & 42.9 & 29.4 & - & 34.4 & 53.5 & 65.1 \\ 
    VL-only  & 32.4 & 27.7 & 34.4 & 18.3 & 31.4 & 35.7 & 47.5 & 58.3 \\
    \approach{} & 80.5 & 69.1 & 49.8 & 38.6 & 39.1 & 46.3 & 57.0 & 66.9 \\
    \bottomrule
  \end{tabular}
  \caption{Ablation study on sing-task baselines of direct evaluation results on ImageNet series and zero-/few-shot evaluation on 12-dataset benchmark.}
  \label{tab-ablation-overall}
\end{table}

\noindent \textbf{Ablation: single-task baseline} Here we compare the proposed approach with two single task baselines, supervised-only (Eqn.~\eqref{eq:supervise}) and VL-only (Eqn.~\eqref{eq:vl-contrastive}) on zero-/few-shot classification on IN-related and 12-dataset benchmarks. From Table~\ref{tab-ablation-overall}, our approach could perform competitively or significantly better than two single-task baselines, which indicates that our deep fusion approach could well absorb the strengths of these two tasks and outperform both.

\begin{table}[t]
  \centering
  \small
  \addtolength{\tabcolsep}{0.7pt}
  \begin{tabular}{c|ccc|cccc|c}
    \toprule
    \multicolumn{1}{c}{} & \multicolumn{3}{c}{Visual Encoder}  & \multicolumn{4}{c}{ImageNet-Related} & 
    \multicolumn{1}{c}{K. 12-dataset} \\
  Method & input & \#param. & FLOPs & IN & IN-V2 & IN-R & IN-S & {Average} \\
    \hline
    CLIP-ViT-B/16~\cite{radford2021clip} & $224^2$ & 86M & 18.9G & 68.6 & 61.9 & 76.4 & 46.6 & 68.8 \\
    CLIP-ViT-L/14~\cite{radford2021clip} & $336^2$ & 307M & 190.7G & 76.2 & 70.1 & 88.9 & 60.2 & 75.5 \\
    ALIGN-EffNet-L2~\cite{jia2021align} & $360^2$ & 480M & 92.2G & 76.4 & 70.1 & 92.2 & - & -\\  
    \hline
    Sup-only & $224^2$ & 88M & 15.4G & 82.6 & 73.1 & 56.5 & 42.0 & -\\
    VL-only  & $224^2$ & 88M & 15.4G & 60.1  & 55.1 & 73.6 & 51.1 & 65.6 \\
    \approach{} & $224^2$ & 88M & 15.4G & 82.9 & 73.7 & 76.7 & 59.8 & 70.6 \\
    \bottomrule
  \end{tabular}
  \caption{System-level comparison with the state-of-the-art approaches (CLIP and ALIGN) of direct evaluation results (top-1 accuracy) on ImageNet series (IN, IN-V2, IN-Rendition, IN-Sketch) and zero-shot evaluation results (top-1 accuracy) on the Kornblith 12-dataset benchmark. }
  \label{tab-overall-zeroshot}
\end{table}

\subsection{System-level Comparison}
\noindent \textbf{Setup}~~We adopt Laion-400M~\cite{schuhmann2021laion400m} as the image-caption dataset and ImageNet-22K~\cite{deng2009imagenet} as the image-label dataset. Both these datasets are open-sourced, thus all the experiments in our paper is reproducible for the whole community. In all experiments, we adopt a base version of Swin Transformer with input size 224$\times$224 and window size 7 as the visual encoder, and a base version of RoBERTa~\cite{liu2019roberta} model as the text encoder.  More implementation details are shown in Appendix~\ref{appd:details}.

\noindent \textbf{Direct/Zero-shot Classification}~~We evaluate our model on ImageNet-1K~\cite{deng2009imagenet} with several variants of validation sets~\cite{imagenetv2,hendrycks2021many,wang2019learningSketch} and Kornblith 12-dataset benchmark~\cite{imagnettransfer}. 
Table~\ref{tab-overall-zeroshot} presents the comparisons to the state-of-the-art models, CLIP~\cite{radford2021clip} and ALIGN~\cite{jia2021align}. Compared to CLIP ViT-B/16 model which has similar number of parameters and FLOPs, our \approach{} gets over 12\% improvements on IN, IN-V2 and IN-S and obtains slightly better performance on IN-R. Achieving better performance on IN and IN-V2 is reasonable because IN-22K contains plenty of images in IN-1K. The results on ImageNet variants are higher or at least on-par with the largest model of CLIP and ALIGN, which has either ten times of FLOPs or five times of parameters. Besides, we find that a unified combination of image classification and image-text alignment empowers the model to learn knowledge from both tasks and datasets, better than strong supervised-22K and VL-only baseline on all the ImageNet related validation sets.

When evaluating the models on Kornblith 12-dataset classification benchmark~\cite{imagnettransfer} which covers a wide range of visual recognition domains, our model outperforms CLIP ViT-B/16 model by 1.8\% on averaged accuracy of 12 datasets. On some fine-grained datasets, like DTD (Texture) and Flowers102 (Flowers), our model is better than the biggest CLIP ViT-L/14 (336$\uparrow$) model. And on more general datasets, like CIFAR100 and Caltech101, our models can still achieve on-par performance with a way larger CLIP model. Compared to VL-only baseline, our model can achieve a huge gain of 5.0\% on average, due to a unified contrastive learning during pre-training. The detailed results are shown in Appendix~\ref{appd:results-details}.

\begin{figure}[t]
\begin{minipage}[c]{0.55\textwidth}
    \centering
    \includegraphics[width=0.95\linewidth]{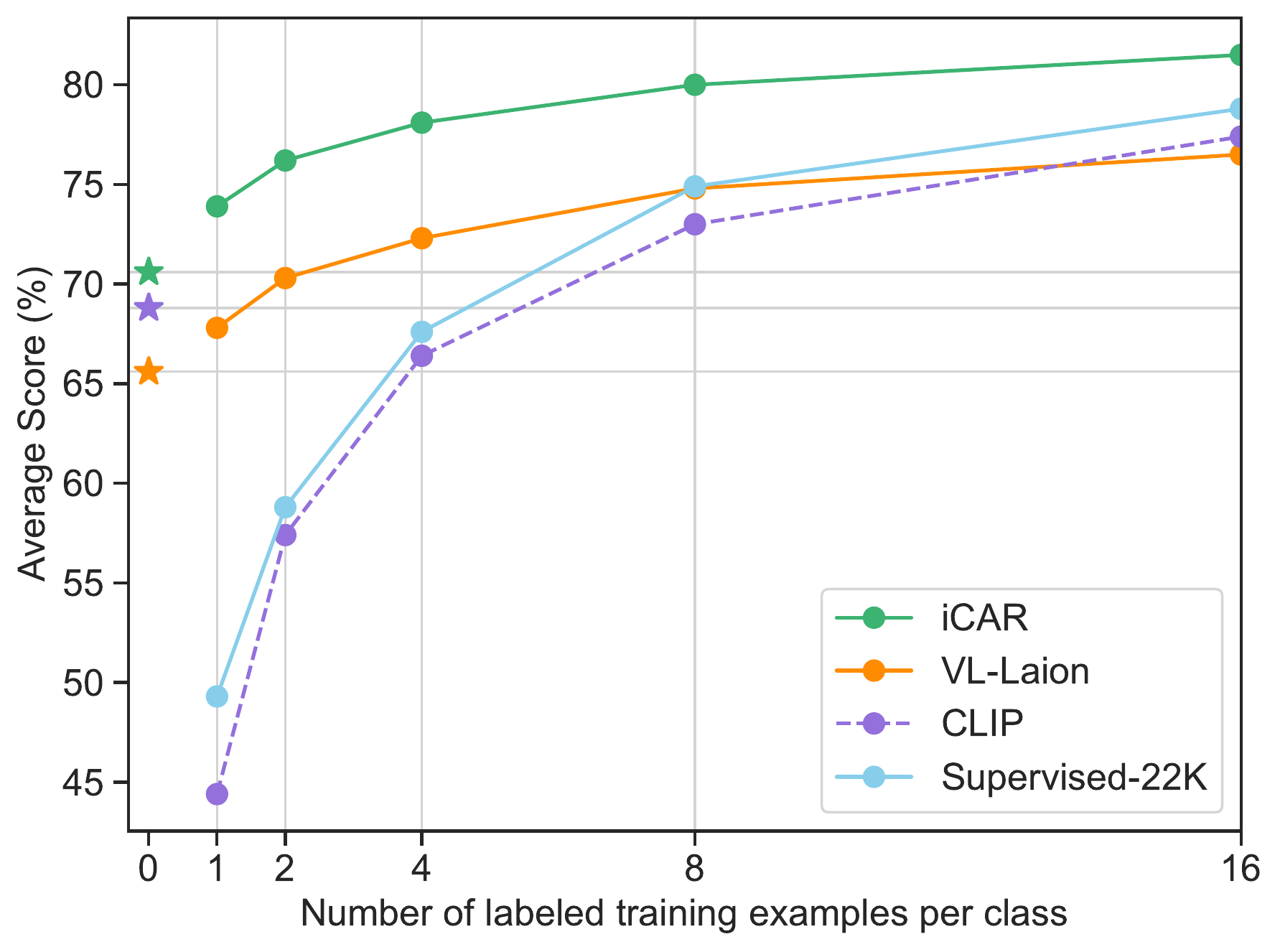}
\end{minipage}
\begin{minipage}[c]{0.4\textwidth}
    \caption{Major comparison with the state-of-the-art approach CLIP of few-shot classification (top-1 accuracy) on the Kornblith  12-dataset benchmark. $\star$ denotes the zero-shot performance of different approaches. Results of CLIP-ViT-B/16 on few-shot classification are reproduced using released model.}
    \label{fig:overall-fewshot}
\end{minipage}
\end{figure}

\noindent \textbf{Few-shot Classification}~~We compare our model with CLIP under the few-shot classification scenario with the visual encoder frozen in Figure~\ref{fig:overall-fewshot}. We follow~\cite{radford2021clip} to train a linear probe CLIP with a careful grid search on hyper-parameters. We notice that in CLIP, the performance of few-shot classification cannot catch up with that of zero-shot classification, unless more than 4 examples per class are given. We assume that it is because the number of training samples is not enough for training a randomly initialized classifier. This situation can be alleviated by fine-tuning with the pre-trained text encoder which serves as a better initialization, and this method closes the gap between pre-training and fine-tuning. We evaluate our method on Kornblith 12-dataset benchmark. We run every experiments three times and the averaged results are reported. 

When only one example per class is given, by utilizing text encoder as the classifier, our \approach{} achieve 73.9\% on 12-dataset average, surpassing the CLIP model by 29.5\%. And our model earns 3.3\% better than our zero-shot baseline which already has a strong performance. Even trained with 16 examples per class, our model can still surpass CLIP method by 4.1\%. Compared to supervised-only model and visual-linguistic only model, our unified contrastive learning pre-trained model is 24.6\% and 6.1\% better respectively under one-shot setting, and the advantage keeps to 16 shots learning with still 2.7\% and 5.0\% better. 

We also evaluate our model on zero-shot cross-modal retrieval benchmarks of Flickr-30K (1K test split) and MSCOCO (5K test split). Our approach achieves highly competitive results with CLIP. We list our results in Appendix~\ref{appd:results-retrieval-details}.

\subsection{Fine-tuning on Downstream Tasks}\label{exp:finetune}

\begin{table}[t]
  \small
  \centering
  \addtolength{\tabcolsep}{2.5pt}
  \renewcommand{\arraystretch}{0.95}
  \begin{tabular}{c| @{\extracolsep{\fill}} cccc}
    \toprule
    & \multicolumn{4}{c}{Evaluation Dataset (val mIoU)} \\
    Method (zero-shot) & ADE-20K& Cityscapes & VOC & COCO Stuff\\
    \hline
    CLIP  & 1.7 & 0.8 & 12.6 & 1.0 \\
    VL-Laion & 8.9 & 6.7 & 22.2 & 2.1 \\
    \approach{} & 9.7 & 10.0 & 21.3 & 2.1 \\
    \midrule
    \midrule
    & \ Ft. Dataset & \multicolumn{3}{c}{Open-vocabulary Dataset} \\
    Method & ADE-20K& Cityscapes & VOC & COCO Stuff \\
    \hline
    \demph{Sup-only-Ft.} & \demph{52.1} & \demph{-} & \demph{-} & \demph{-} \\ 
    \demph{VL-only-Ft.} & \demph{51.9} & \demph{-} & \demph{-} & \demph{-} \\ 
    \approach{}-Ft. (visual enc. only) & 52.6 & - & - & - \\
    \approach{}-Ft. & 52.5 & 53.8 & 47.7 & 14.7 \\
    \bottomrule
  \end{tabular}
  \caption{Results of zero-shot evaluation (val mIoU) on four semantic segmentation benchmarks, and open-vocabulary performance (val mIoU) on three of four datasets (Cityscapes, VOC, COCO Stuff) when performing the text encoder included fine-tuning on ADE20K with MaskFormer as default framework. \approach{}-Ft. (visual enc. only) denotes fine-tuning only the visual encoder of the pre-trained model. \approach{}-Ft. denotes further integrating text encoder as a classifier during fine-tuning.}
  \label{tab-ft-segmentation}
\end{table}

\noindent \textbf{Semantic Segmentation}~~We conduct the experiment mainly on the widely-used ADE-20K~\cite{zhou2018semantic} dataset, including 150 categories and 25K images. We utilize MaskFormer~\cite{cheng2021maskformer} as our base framework and adopt its default training recipe except for setting window size to 7. MaskFormer is a two-stage framework for segmentation and therefore suitable for our situation. We replace the classifier with the pre-trained text encoder, i.e. we generate the classifier weight on-the-fly by feeding each class name and a prompt into the text encoder. For generated masks which should be matched to the empty set $\varnothing$, we add a special category - "background". We fine-tune our pre-trained model on ADE-20K dataset and conduct open-vocabulary semantic segmentation on Pascal VOC~\cite{PASCAL}, Cityscapes~\cite{cordts2016cityscapes} and COCO Stuff~\cite{lin2014coco}. We compare our approach with zero-shot baselines and the baseline fine-tuning visual backbone only, which preserves original multi-way classifier and has no ability for open-vocabulary segmentation. For zero-shot baselines, we extract the feature map after the last stage and conduct a point-wise classification following an interpolation to the original image size.

Table~\ref{tab-ft-segmentation} shows single scale testing results of validation mIoU in different datasets. We find that fine-tuning with the text encoder can achieve an on-par result compared to the baseline which directly uses 151-way classification on ADE-20K (52.5 v.s. 52.6). Furthermore, our method shows surprisingly good transferability on other open-vocabulary segmentation datasets, achieving 47.7 mIoU on Pascal VOC, 53.8 mIoU on Cityscapes, 14.7 mIoU on COCO Stuff. Compared to the best zero-shot baseline, our method incurs huge improvements, with +26.4, +43.8, +12.5 on these three datasets, respectively, showing the surprisingly good open vocabulary capability of our model.

\begin{table}[t]
  \centering
  \small
  \addtolength{\tabcolsep}{2pt}
  \begin{tabular}{c| @{\extracolsep{\fill}} cccc}
    \toprule
    \centering
     & Ft. Dataset & \multicolumn{3}{c}{Open-vocabulary Dataset} \\
     Method & \ LVIS (Rare) & COCO & Objects365 & VOC \\
    \hline
    \demph{Sup-only-Ft.} & \demph{35.9 (25.2)} & \demph{-} & \demph{-} & \demph{-} \\
    \demph{VL-only-Ft.} & \demph{36.3 (25.9)} & \demph{-} & \demph{-} & \demph{-} \\
    \approach{}-Ft. (visual enc. only) & 37.9 (29.0) & - & - & - \\
    \approach{}-Ft. & 38.5 (30.8) & 41.2 & 19.7 & 76.4 \\
    \bottomrule
  \end{tabular}
  \caption{Results of open-vocabulary performance (box AP on validation set) on three object detection datasets (COCO, Objects365, Pascal VOC) when performing the text encoder included fine-tuning on LVIS with Faster R-CNN as the default framework.}
  \label{tab-ft-detection}
\end{table}

\noindent \textbf{Object Detection}~~We conduct the object detection experiment on LVIS v1~\cite{gupta2019lvis} with Faster R-CNN~\cite{ren2015faster} and FPN~\cite{he2017fpn} as framework. LVIS includes over 1200 object categories with an unbalanced distribution. Based on the frequency, all the categories are splitted into three sets: f(requency),c(ommon),r(are). Similar to the setting in semantic segmentation, we replace bounding box classification head with the pre-trained text encoder. But we do not add in a special ``background" category, instead we keep it as a pure trainable embedding. We fine-tune our model for 2x schedule (24 epochs) on LVIS dataset, with multi-scale training (shorter size between 480 and 800) and a cosine learning rate schedule mostly following the recipe from~\cite{Swin}. 

After fine-tuning, we conduct open-vocabulary detection on COCO~\cite{lin2014coco}, Objects365~\cite{Shao_2019_ICCV} and Pascal VOC~\cite{PASCAL}. We test with a single scale input (1333, 800) and report box mAP on validation set, except that for VOC, we report box AP50 and use a default (1000, 600) input.

Table~\ref{tab-ft-detection} shows the comparison results of our approach with baselines on four datasets.  When fine-tuning on LVIS with the text encoder as visual classifier, our approach obtains 41.2 box mAP on COCO without any annotations, which is only 7.4 points lower than fully supervised fine-tuning baseline with IN-22K pre-trained model as initialization. Besides, we also achieve 19.7 mAP on O365 and 76.4 AP50 on VOC. In addition, our approach gains +0.6 mAP better on LVIS (38.5 mAP v.s. 37.9 mAP) than baseline, which has a 1204-way classifier. 

\begin{table}[t]
  \small
  \centering
    \addtolength{\tabcolsep}{1.8pt}
    \renewcommand{\arraystretch}{0.95}
  \begin{tabular}{c| @{\extracolsep{\fill}} cccc}
    \toprule
    & \multicolumn{4}{c}{Evaluation Dataset (top-1 accuracy)} \\
    Method (zero-shot) & K400 & K600 & UCF101 & HMDB51 \\
    \hline
    CLIP  & 51.3 & 47.7 & 67.4 & 46.1 \\
    VL-Laion & 40.5 & 37.5 & 56.5 & 36.0 \\
    \approach{} & 46.4 & 43.7 & 61.4 & 41.2 \\
    \midrule\midrule
    & \ Ft. Dataset & \multicolumn{3}{c}{Open-vocabulary Dataset} \\
    Method & K400 & K600 & UCF101 & HMDB51 \\
    \hline
    \demph{Sup-only-Ft.}  & \demph{82.7} & \demph{-} & \demph{-} & \demph{-} \\ 
    \demph{VL-only-Ft.}  & \demph{82.1} & \demph{-} & \demph{-} & \demph{-} \\ 
    \approach{}-Ft. (visual enc. only) & 83.1 & - & - & - \\
    \approach{}-Ft. & 83.2 & 59.5 & 73.0 & 45.4 \\
    \bottomrule
  \end{tabular}
  \caption{Results of zero-shot evaluation (top-1 accuracy) on four video action recognition benchmarks, and open-vocabulary evaluation (top-1 accuracy) on three of four datasets (Kinetics-600, UCF101, HMDB51) when performing the text encoder included fine-tuning on Kinetics-400 with Video-Swin as default framework. The zero-shot result of CLIP on UCF101 is reproduced with released model.}
  \label{tab-ft-video}
\end{table}

\noindent \textbf{Video Action Recognition}~~We also evaluate our approach on the video action recognition task, following the same recipe in Video Swin Transformer~\cite{liu2021videoswin} except that we integrate the pre-trained text encoder as our classifier. We fine-tune our model on Kinetics-400 (K400)~\cite{kay2017kinetics} dataset for 30 epochs and conduct open-vocabulary recognition on three other datasets, Kinetics-600 (K600)~\cite{kay2017kinetics}, UCF101~\cite{ucf101} and HMDB51~\cite{hmdb51}. K400 consists of 400 human action categories and is widely used in video action recognition. The overlap videos between K600 validation set and K400 training set are carefully removed. We also compare with several zero-shot baselines following the instruction mentioned in CLIP~\cite{radford2021clip}.

Table~\ref{tab-ft-video} presents top-1 accuracy of each method on four datasets. Our model fine-tuned on K400 earns 59.5\% top-1 accuracy on K600, surpassing our zero-shot baseline for +15.8\%. To deeply understand this result, we disentangle the good performance on K600, and create a new split named Kinetics-232 from the validation set of K600, containing the classes which are not overlapped with the ones in K400. We discover that our model after fine-tuning could still perform relatively well on this split full of unseen classes in fine-tuning, with only 11.4\% accuracy dropped compared to zero-shot baseline. With benefits from the other 368 classes, our model after fine-tuning reaches higher on K600.  On the other two datasets, our model reaches 73.0\% and 45.4\% on UCF101 and HMDB51 respectively, with +11.6\% and +4.2\% better than zero-shot baseline, which reveals the open-vocabulary generalization ability of our approach. And our approach performs equally with standard fine-tuning approach on K400 (83.2\% v.s. 83.1\%), which loads the visual encoder only and adds a multi-way parametric classifier for classification.

\section{Conclusion}
\label{sec:conclusion}

In this paper, we present a new approach for bridging the image classification task with the image-text alignment task, from the perspective of training loss, classifier type and label granularity. The deep unification could help the model benefit from both tasks and achieve significantly better performance than the single-task baselines or the simple joint training baseline with separate task heads. The effectiveness of the proposed approach is verified on a wide range of tasks, such as zero-/few-shot classification on ImageNet related benchmarks and Kornblith 12-dataset benchmarks, and fine-tuning on three representative downstream tasks of both close-set and open-vocabulary scenarios.

% \clearpage
% ---- Bibliography ----
%
% BibTeX users should specify bibliography style 'splncs04'.
% References will then be sorted and formatted in the correct style.
%
\bibliographystyle{splncs04}
\bibliography{egbib}

\begin{thebibliography}{10}
\providecommand{\url}[1]{\texttt{#1}}
\providecommand{\urlprefix}{URL }
\providecommand{\doi}[1]{https://doi.org/#1}

\bibitem{carion2020detr}
Carion, N., Massa, F., Synnaeve, G., Usunier, N., Kirillov, A., Zagoruyko, S.:
  End-to-end object detection with transformers. In: European Conference on
  Computer Vision. pp. 213--229. Springer (2020)

\bibitem{carreira2017i3d}
Carreira, J., Zisserman, A.: Quo vadis, action recognition? a new model and the
  kinetics dataset. In: proceedings of the IEEE Conference on Computer Vision
  and Pattern Recognition. pp. 6299--6308 (2017)

\bibitem{changpinyo2021cc12m}
Changpinyo, S., Sharma, P., Ding, N., Soricut, R.: {Conceptual 12M}: Pushing
  web-scale image-text pre-training to recognize long-tail visual concepts. In:
  CVPR (2021)

\bibitem{chen2017rethinking}
Chen, L.C., Papandreou, G., Schroff, F., Adam, H.: Rethinking atrous
  convolution for semantic image segmentation. arXiv preprint arXiv:1706.05587
  (2017)

\bibitem{cheng2021maskformer}
Cheng, B., Schwing, A.G., Kirillov, A.: Per-pixel classification is not all you
  need for semantic segmentation. arXiv  (2021)

\bibitem{cordts2016cityscapes}
Cordts, M., Omran, M., Ramos, S., Rehfeld, T., Enzweiler, M., Benenson, R.,
  Franke, U., Roth, S., Schiele, B.: The cityscapes dataset for semantic urban
  scene understanding. In: Proceedings of the IEEE conference on computer
  vision and pattern recognition. pp. 3213--3223 (2016)

\bibitem{cubuk2020randaugment}
Cubuk, E.D., Zoph, B., Shlens, J., Le, Q.V.: Randaugment: Practical automated
  data augmentation with a reduced search space. In: Proceedings of the
  IEEE/CVF Conference on Computer Vision and Pattern Recognition Workshops. pp.
  702--703 (2020)

\bibitem{deng2009imagenet}
Deng, J., Dong, W., Socher, R., Li, L.J., Li, K., Fei-Fei, L.: Imagenet: A
  large-scale hierarchical image database. In: 2009 IEEE conference on computer
  vision and pattern recognition. pp. 248--255. Ieee (2009)

\bibitem{desai2021virtex}
Desai, K., Johnson, J.: Virtex: Learning visual representations from textual
  annotations. In: Proceedings of the IEEE/CVF Conference on Computer Vision
  and Pattern Recognition. pp. 11162--11173 (2021)

\bibitem{dosovitskiy2020vit}
Dosovitskiy, A., Beyer, L., Kolesnikov, A., Weissenborn, D., Zhai, X.,
  Unterthiner, T., Dehghani, M., Minderer, M., Heigold, G., Gelly, S.,
  Uszkoreit, J., Houlsby, N.: An image is worth 16x16 words: Transformers for
  image recognition at scale. In: International Conference on Learning
  Representations (2021), \url{https://openreview.net/forum?id=YicbFdNTTy}

\bibitem{PASCAL}
Everingham, M., Gool, L., Williams, C.K., Winn, J., Zisserman, A.: The pascal
  visual object classes (voc) challenge. Int. J. Comput. Vision
  \textbf{88}(2),  303–338 (Jun 2010). \doi{10.1007/s11263-009-0275-4},
  \url{https://doi.org/10.1007/s11263-009-0275-4}

\bibitem{feichtenhofer2018slowfast}
Feichtenhofer, C., Fan, H., Malik, J., He, K.: Slowfast networks for video
  recognition. In: Proceedings of the IEEE international conference on computer
  vision. pp. 6202--6211 (2019)

\bibitem{nips13devise}
Frome, A., Corrado, G.S., Shlens, J., Bengio, S., Dean, J., Ranzato, M.A.,
  Mikolov, T.: Devise: A deep visual-semantic embedding model. In: Burges,
  C.J.C., Bottou, L., Welling, M., Ghahramani, Z., Weinberger, K.Q. (eds.)
  Advances in Neural Information Processing Systems. vol.~26. Curran
  Associates, Inc. (2013),
  \url{https://proceedings.neurips.cc/paper/2013/file/7cce53cf90577442771720a370c3c723-Paper.pdf}

\bibitem{girshick2014rich}
Girshick, R., Donahue, J., Darrell, T., Malik, J.: Rich feature hierarchies for
  accurate object detection and semantic segmentation (2014)

\bibitem{gomez2017self}
Gomez, L., Patel, Y., Rusi{\~n}ol, M., Karatzas, D., Jawahar, C.:
  Self-supervised learning of visual features through embedding images into
  text topic spaces. In: Proceedings of the IEEE Conference on Computer Vision
  and Pattern Recognition. pp. 4230--4239 (2017)

\bibitem{gupta2019lvis}
Gupta, A., Dollar, P., Girshick, R.: Lvis: A dataset for large vocabulary
  instance segmentation. In: Proceedings of the IEEE/CVF Conference on Computer
  Vision and Pattern Recognition. pp. 5356--5364 (2019)

\bibitem{he2017mask}
He, K., Gkioxari, G., Doll{\'a}r, P., Girshick, R.: Mask r-cnn. In: Proceedings
  of the IEEE international conference on computer vision. pp. 2961--2969
  (2017)

\bibitem{he2015resnet}
He, K., Zhang, X., Ren, S., Sun, J.: Deep residual learning for image
  recognition. In: Proceedings of the IEEE conference on computer vision and
  pattern recognition. pp. 770--778 (2016)

\bibitem{hendrycks2021many}
Hendrycks, D., Basart, S., Mu, N., Kadavath, S., Wang, F., Dorundo, E., Desai,
  R., Zhu, T., Parajuli, S., Guo, M., Song, D., Steinhardt, J., Gilmer, J.: The
  many faces of robustness: A critical analysis of out-of-distribution
  generalization. ICCV  (2021)

\bibitem{huang2016deep}
Huang, G., Sun, Y., Liu, Z., Sedra, D., Weinberger, K.Q.: Deep networks with
  stochastic depth. In: European conference on computer vision. pp. 646--661.
  Springer (2016)

\bibitem{jia2021align}
Jia, C., Yang, Y., Xia, Y., Chen, Y.T., Parekh, Z., Pham, H., Le, Q.V., Sung,
  Y., Li, Z., Duerig, T.: Scaling up visual and vision-language representation
  learning with noisy text supervision. arXiv preprint arXiv:2102.05918  (2021)

\bibitem{joulin2016first-flickr-vl}
Joulin, A., Van Der~Maaten, L., Jabri, A., Vasilache, N.: Learning visual
  features from large weakly supervised data. In: European Conference on
  Computer Vision. pp. 67--84. Springer (2016)

\bibitem{kay2017kinetics}
Kay, W., Carreira, J., Simonyan, K., Zhang, B., Hillier, C., Vijayanarasimhan,
  S., Viola, F., Green, T., Back, T., Natsev, P., et~al.: The kinetics human
  action video dataset. arXiv preprint arXiv:1705.06950  (2017)

\bibitem{khosla2020supervised}
Khosla, P., Teterwak, P., Wang, C., Sarna, A., Tian, Y., Isola, P., Maschinot,
  A., Liu, C., Krishnan, D.: Supervised contrastive learning. arXiv preprint
  arXiv:2004.11362  (2020)

\bibitem{kingma2014adam}
Kingma, D.P., Ba, J.: Adam: A method for stochastic optimization. arXiv
  preprint arXiv:1412.6980  (2014)

\bibitem{imagnettransfer}
Kornblith, S., Shlens, J., Le, Q.V.: Do better imagenet models transfer better?
  In: 2019 IEEE/CVF Conference on Computer Vision and Pattern Recognition
  (CVPR). pp. 2656--2666 (2019). \doi{10.1109/CVPR.2019.00277}

\bibitem{StanfordCars}
Krause, J., Stark, M., Deng, J., Fei-Fei, L.: 3d object representations for
  fine-grained categorization. In: 2013 IEEE International Conference on
  Computer Vision Workshops. pp. 554--561 (2013). \doi{10.1109/ICCVW.2013.77}

\bibitem{krizhevsky2012alexnet}
Krizhevsky, A., Sutskever, I., Hinton, G.E.: Imagenet classification with deep
  convolutional neural networks. In: Advances in neural information processing
  systems. pp. 1097--1105 (2012)

\bibitem{hmdb51}
Kuehne, H., Jhuang, H., Garrote, E., Poggio, T., Serre, T.: {HMDB}: a large
  video database for human motion recognition. In: Proceedings of the
  International Conference on Computer Vision (ICCV) (2011)

\bibitem{he2017fpn}
Lin, T.Y., Dollar, P., Girshick, R., He, K., Hariharan, B., Belongie, S.:
  Feature pyramid networks for object detection. In: The IEEE Conference on
  Computer Vision and Pattern Recognition (CVPR) (July 2017)

\bibitem{lin2014coco}
Lin, T.Y., Maire, M., Belongie, S., Hays, J., Perona, P., Ramanan, D.,
  Doll{\'a}r, P., Zitnick, C.L.: Microsoft coco: Common objects in context. In:
  European conference on computer vision. pp. 740--755. Springer (2014)

\bibitem{liu2019roberta}
Liu, Y., Ott, M., Goyal, N., Du, J., Joshi, M., Chen, D., Levy, O., Lewis, M.,
  Zettlemoyer, L., Stoyanov, V.: Roberta: A robustly optimized bert pretraining
  approach. arXiv preprint arXiv:1907.11692  (2019)

\bibitem{Swin}
Liu, Z., Lin, Y., Cao, Y., Hu, H., Wei, Y., Zhang, Z., Lin, S., Guo, B.: Swin
  transformer: Hierarchical vision transformer using shifted windows. CoRR
  \textbf{abs/2103.14030} (2021), \url{https://arxiv.org/abs/2103.14030}

\bibitem{liu2021videoswin}
Liu, Z., Ning, J., Cao, Y., Wei, Y., Zhang, Z., Lin, S., Hu, H.: Video swin
  transformer. arXiv preprint arXiv:2106.13230  (2021)

\bibitem{long2015fully}
Long, J., Shelhamer, E., Darrell, T.: Fully convolutional networks for semantic
  segmentation (2015)

\bibitem{aircraft}
Maji, S., Kannala, J., Rahtu, E., Blaschko, M., Vedaldi, A.: Fine-grained
  visual classification of aircraft. Tech. rep. (2013)

\bibitem{miller-1994-wordnet}
Miller, G.A.: {W}ord{N}et: A lexical database for {E}nglish. In: {H}uman
  {L}anguage {T}echnology: Proceedings of a Workshop held at {P}lainsboro,
  {N}ew {J}ersey, {M}arch 8-11, 1994 (1994),
  \url{https://aclanthology.org/H94-1111}

\bibitem{norouzi2013zero}
Norouzi, M., Mikolov, T., Bengio, S., Singer, Y., Shlens, J., Frome, A.,
  Corrado, G.S., Dean, J.: Zero-shot learning by convex combination of semantic
  embeddings. arXiv preprint arXiv:1312.5650  (2013)

\bibitem{qiu2017P3D}
Qiu, Z., Yao, T., Mei, T.: Learning spatio-temporal representation with
  pseudo-3d residual networks. In: proceedings of the IEEE International
  Conference on Computer Vision. pp. 5533--5541 (2017)

\bibitem{radford2021clip}
Radford, A., Kim, J.W., Hallacy, C., Ramesh, A., Goh, G., Agarwal, S., Sastry,
  G., Askell, A., Mishkin, P., Clark, J., Krueger, G., Sutskever, I.: Learning
  transferable visual models from natural language supervision (2021)

\bibitem{imagenetv2}
Recht, B., Roelofs, R., Schmidt, L., Shankar, V.: Do {I}mage{N}et classifiers
  generalize to {I}mage{N}et? In: Chaudhuri, K., Salakhutdinov, R. (eds.)
  Proceedings of the 36th International Conference on Machine Learning.
  Proceedings of Machine Learning Research, vol.~97, pp. 5389--5400. PMLR
  (09--15 Jun 2019), \url{http://proceedings.mlr.press/v97/recht19a.html}

\bibitem{ren2015faster}
Ren, S., He, K., Girshick, R., Sun, J.: Faster r-cnn: Towards real-time object
  detection with region proposal networks. In: Advances in neural information
  processing systems. pp. 91--99 (2015)

\bibitem{sariyildiz2020icmlm}
Sariyildiz, M.B., Perez, J., Larlus, D.: Learning visual representations with
  caption annotations. In: European Conference on Computer Vision (ECCV) (2020)

\bibitem{schuhmann2021laion400m}
Schuhmann, C., Vencu, R., Beaumont, R., Kaczmarczyk, R., Mullis, C., Katta, A.,
  Coombes, T., Jitsev, J., Komatsuzaki, A.: Laion-400m: Open dataset of
  clip-filtered 400 million image-text pairs (2021)

\bibitem{Shao_2019_ICCV}
Shao, S., Li, Z., Zhang, T., Peng, C., Yu, G., Zhang, X., Li, J., Sun, J.:
  Objects365: A large-scale, high-quality dataset for object detection. In:
  Proceedings of the IEEE/CVF International Conference on Computer Vision
  (ICCV) (October 2019)

\bibitem{sharma2018CC3m}
Sharma, P., Ding, N., Goodman, S., Soricut, R.: Conceptual captions: A cleaned,
  hypernymed, image alt-text dataset for automatic image captioning. In:
  Proceedings of ACL (2018)

\bibitem{simonyan2014vgg}
Simonyan, K., Zisserman, A.: Very deep convolutional networks for large-scale
  image recognition. In: International Conference on Learning Representations
  (May 2015)

\bibitem{ucf101}
Soomro, K., Zamir, A.R., Shah, M.: Ucf101: A dataset of 101 human actions
  classes from videos in the wild. CoRR  \textbf{abs/1212.0402} (2012)

\bibitem{szegedy2015googlenet}
Szegedy, C., Liu, W., Jia, Y., Sermanet, P., Reed, S., Anguelov, D., Erhan, D.,
  Vanhoucke, V., Rabinovich, A.: Going deeper with convolutions. In:
  Proceedings of the IEEE conference on computer vision and pattern
  recognition. pp.~1--9 (2015)

\bibitem{tran2015learning}
Tran, D., Bourdev, L., Fergus, R., Torresani, L., Paluri, M.: Learning
  spatiotemporal features with 3d convolutional networks. In: Proceedings of
  the IEEE international conference on computer vision. pp. 4489--4497 (2015)

\bibitem{InfoNCE}
{van den Oord}, A., {Li}, Y., {Vinyals}, O.: {Representation Learning with
  Contrastive Predictive Coding}. arXiv e-prints arXiv:1807.03748 (Jul 2018)

\bibitem{wang2019learningSketch}
Wang, H., Ge, S., Lipton, Z., Xing, E.P.: Learning robust global
  representations by penalizing local predictive power. In: Advances in Neural
  Information Processing Systems. pp. 10506--10518 (2019)

\bibitem{wu2018improving}
Wu, Z., Efros, A.A., Yu, S.X.: Improving generalization via scalable
  neighborhood component analysis. In: Proceedings of the European Conference
  on Computer Vision (ECCV). pp. 685--701 (2018)

\bibitem{xie2017resnext}
Xie, S., Girshick, R., Doll{\'a}r, P., Tu, Z., He, K.: Aggregated residual
  transformations for deep neural networks. In: Proceedings of the IEEE
  Conference on Computer Vision and Pattern Recognition. pp. 1492--1500 (2017)

\bibitem{yao2021filip}
Yao, L., Huang, R., Hou, L., Lu, G., Niu, M., Xu, H., Liang, X., Li, Z., Jiang,
  X., Xu, C.: Filip: Fine-grained interactive language-image pre-training
  (2021)

\bibitem{yin2020DNL}
Yin, M., Yao, Z., Cao, Y., Li, X., Zhang, Z., Lin, S., Hu, H.: Disentangled
  non-local neural networks. In: Proceedings of the European conference on
  computer vision (ECCV) (2020)

\bibitem{young2014flickr}
Young, P., Lai, A., Hodosh, M., Hockenmaier, J.: From image descriptions to
  visual denotations: New similarity metrics for semantic inference over event
  descriptions. Transactions of the Association for Computational Linguistics
  \textbf{2},  67--78 (2014)

\bibitem{zhang2020vl-medical}
Zhang, Y., Jiang, H., Miura, Y., Manning, C.D., Langlotz, C.P.: Contrastive
  learning of medical visual representations from paired images and text. arXiv
  preprint arXiv:2010.00747  (2020)

\bibitem{zhou2018semantic}
Zhou, B., Zhao, H., Puig, X., Xiao, T., Fidler, S., Barriuso, A., Torralba, A.:
  Semantic understanding of scenes through the ade20k dataset. International
  Journal on Computer Vision  (2018)

\end{thebibliography}

\appendix
\section{Implementation Details}
\label{appd:details}
\subsection{System-level Comparison}
\noindent \textbf{Datasets}~~We adopt Laion-400M~\cite{schuhmann2021laion400m} as the image-caption dataset and ImageNet-22K~\cite{deng2009imagenet} as the image-label dataset. Laion-400M is the largest openly available visual-linguistic dataset with 400 million pairs. It extracts image-caption pairs in random web pages crawled between 2014 and 2021 from the Common Crawl web data, and de-noises the dataset by scoring and filtering numerous raw pairs with a threshold using a CLIP ViT-B/32 ~\cite{radford2021clip} model. ImageNet-22K is the full set of ImageNet, containing 14.2 million images and long-tailed 22K classes. Each class is a synset in 
wordnet~\cite{miller-1994-wordnet} with a lemma and a brief description. Both these datasets are open-sourced, thus all the experiments in our paper is reproducible for the whole community. For the Laion-400M, we follow~\cite{radford2021clip} to train a near-duplicate detector via contrastive learning, and remove the duplicate pairs between validation sets and the training set of Laion-400M.

\noindent \textbf{Implementation Details}~~We train iCAR for 100K iterations, with a batch size of 192 per GPU on 64 V100. In each mini batch, we sample 64 images from ImageNet-22K and 128 images from Laion-400M, so the model is trained on classification dataset for 30 epochs and on image-caption dataset for 2 epochs equivalently. We employ an AdamW~\cite{kingma2014adam} optimizer with the learning rate set to 1e-3 and the weight decay set to 0.05, and a cosine learning rate schedule is adopted with a linear warmup for 16.7K iterations. We also add in other regularization, including RandAugment~\cite{cubuk2020randaugment}, stochastic depth~\cite{huang2016deep} of 0.2 and a gradient clipping with a max norm of 5. For supervised training baseline (Sup-only / Supervised-22K), we use the released version of Swin Transformer~\cite{Swin}, which is trained on ImageNet-22K dataset for 90 epochs. For image-text alignment baseline (VL-only / VL-Laion), we train the same architecture as iCAR for 2 epochs on Laion-400M with the same hyper-parameters.

\begin{table*}[ht!]
  \centering
  \footnotesize
  \begin{tabular}{c|ccccccccccccc}
    \toprule
    Methods &\rotatebox{75}{Food101} &\rotatebox{75}{CIFAR10} &\rotatebox{75}{CIFAR100} &\rotatebox{75}{Birdsnap} &\rotatebox{75}{SUN397} &\rotatebox{75}{Stanford Cars} &\rotatebox{75}{FGVC Aircraft} &\rotatebox{75}{VOC2007} &\rotatebox{75}{DTD} &\rotatebox{75}{Oxford Pets} &\rotatebox{75}{Caltech101} &\rotatebox{75}{Flowers102} &\rotatebox{75}{\textbf{Average}}\\
    \midrule
    CLIP-ViT-B/16 &89.2 &91.6 &68.7 &39.1 &65.2 &65.6 &27.1 &83.9 &46.0 &88.9 &89.3 &70.4 &68.8\\
    CLIP-ViT-L/14 &92.9 &96.2 &77.9 &48.3 &67.7 &77.3 &36.1 &84.1 &55.3 &93.5 &92.6 &78.7 &75.1\\  
    CLIP-ViT-L/14 (336$\uparrow$) &93.8 &95.7 &77.5 &49.5 &68.4 &78.8 &37.2 &84.3 &55.7 &93.5 &92.8 &78.3 &75.5 \\  
    \hline
    VL-Laion  &77.8 &92.4 &68.1 &29.1 &58.0 &65.4 &7.1 &83.9 &62.1 &81.0 &90.0 &71.7 &65.6\\
    iCAR &82.7 &94.8 &78.4 &48.5 &62.9 &63.1 &8.4 &84.5 &62.9 &87.9 &92.1 &81.3 &70.6\\
    \bottomrule
  \end{tabular}
  \caption{Detailed comparisons of zero-shot classification with the state-of-the-art approach CLIP on Kornblith 12-dataset classification benchmark~\cite{imagnettransfer}.}
  \label{tab-zs-details}
\end{table*}
\section{Detailed Results}

\subsection{Zero-shot classification}
\label{appd:results-details}
We compare iCAR with our image-text alignment baseline (VL-Laion) and the state-of-the-art approach CLIP~\cite{radford2021clip} on Kornblith 12-dataset benchmark~\cite{imagnettransfer}. Table~\ref{tab-zs-details} presents the detailed results on each dataset. Compared to image-text alignment baseline, iCAR reaches higher score in 11 out of 12 datasets except Stanford Cars dataset~\cite{StanfordCars} and the average improvement is 5.0\%. And with the help of a unified contrastive learning scheme combining supervised dataset and visual-linguistic dataset, our model could generally perform better than the state-of-the-art CLIP ViT-B/16 approach. The main performance gap between iCAR and CLIP methods is on FGVC Aircraft dataset~\cite{aircraft} and we guess that some variants of the aircraft are a little old and not included in Laion-400M dataset~\cite{schuhmann2021laion400m} which is collected from the websites later than 2014.

\subsection{Zero-shot retrieval} 
\label{appd:results-retrieval-details}
We evaluate iCAR on zero-shot cross-modal retrieval benchmarks of Flickr-30K~\cite{young2014flickr} (1K test split) and MSCOCO~\cite{lin2014coco} (5K test split). Here we follow CLIP~\cite{yao2021filip} to use the similar prompt engineering in this task. Table~\ref{tab-retrieval-major} shows our results and our iCAR achieves comparable results with CLIP on image-to-text retrieval task, surpasses CLIP on text-to-image retrieval task of MSCOCO, and remains a performance gap on Flickr-30K.
\begin{table*}[]
  \centering
  \small
  \begin{tabular}{ccccccccccccc}
    \toprule
     & \multicolumn{6}{c}{Text Retrieval} & \multicolumn{6}{c}{Image Retrieval} \\
    &  \multicolumn{3}{c}{Flickr} &  \multicolumn{3}{c}{MSCOCO}&  \multicolumn{3}{c}{Flickr} &  \multicolumn{3}{c}{MSCOCO} \\
    \cmidrule(lr){2-4} \cmidrule(lr){5-7} \cmidrule(lr){8-10} \cmidrule(lr){11-13}
    Methods & R@1 & R@5 & R@10 &R@1 & R@5 & R@10 &R@1 & R@5 & R@10 &R@1 & R@5 & R@10  \\
    \midrule
    CLIP-ViT-B/16~\cite{radford2021clip} &82.1 &96.6 &99.0 &52.4 &76.7 &84.6 &62.2 &85.7 &91.9 &33.1 &58.4 &69.0\\
    \hline
    VL-Laion &78.5 &95.3 &97.8 &51.7 &76.8 &85.1 &61.3 &85.4 &91.5 &35.6 &61.0 &71.9 \\
    iCAR &81.5 &95.8 &98.5 &52.1 &76.4 &85.3 &59.9 &85.5 &91.2 &35.6 &61.1 &71.6 \\
    \bottomrule
  \end{tabular}
  \caption{Comparisons of our approach and state-of-the-art methods on zero-shot cross-modal retrieval performance on Flickr-30K and MSCOCO.}
  \label{tab-retrieval-major}
\end{table*}

\subsection{Few-shot classification}
\begin{figure*}
    \centering
    \includegraphics[width=\linewidth]{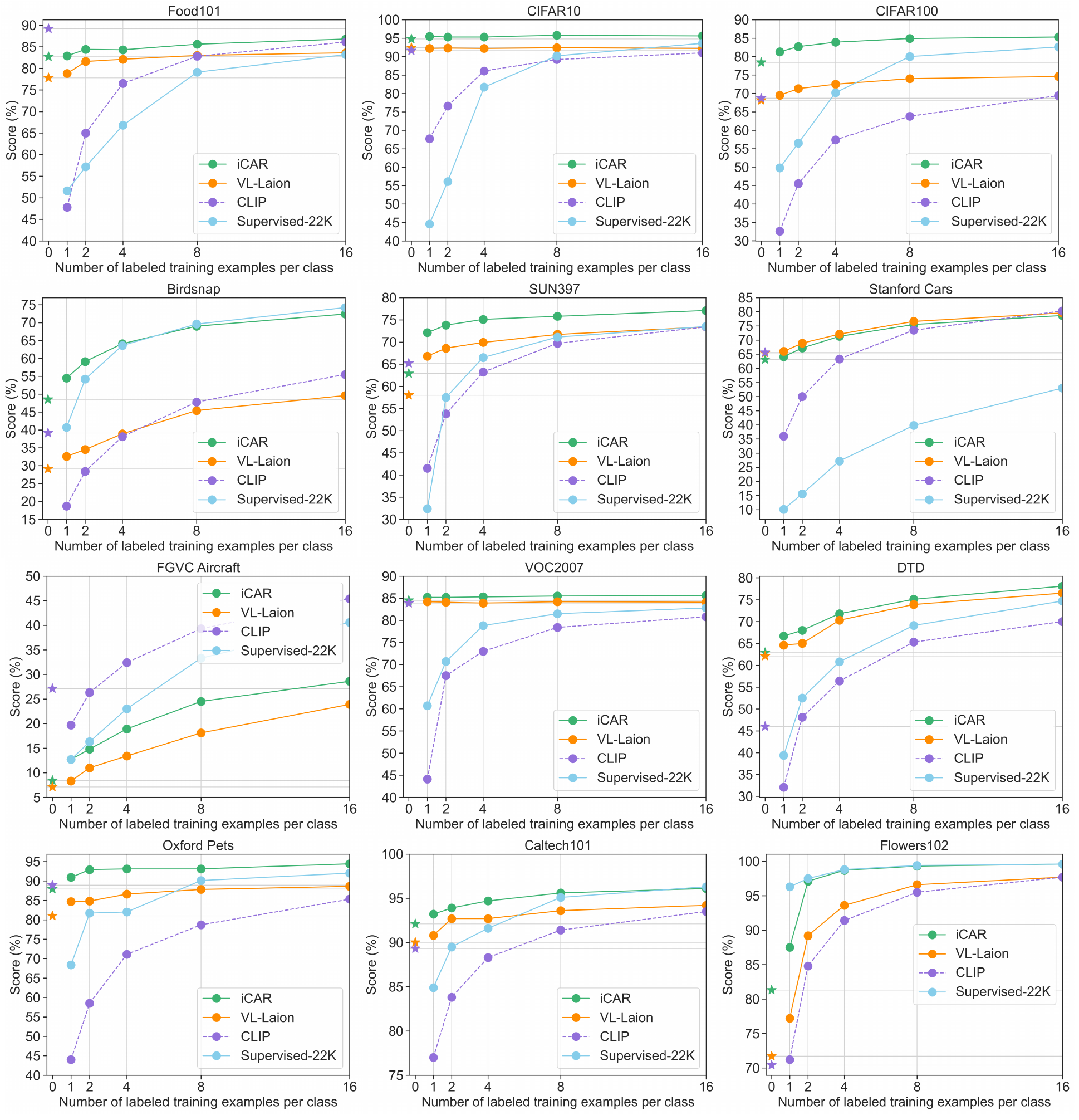}
    \caption{Detailed comparisons with the state-of-the-art approach CLIP of few-shot classification (top-1 accuracy) on each dataset of Kornblith  12-dataset benchmark. $\star$ denotes the zero-shot performance of different approaches. Results of CLIP ViT-B/16 on few-shot classification are reproduced using the released model.}
    \label{fig:fw-details}
\end{figure*}

Figure~\ref{fig:fw-details} shows few-shot classification comparison on each dataset of Kornblith 12-dataset benchmark. With the pre-trained text encoder as an initialized meta classifier, iCAR achieves higher performance under any-shot setting than zero-shot classification. It proves that the gap between zero-shot and few-shot learning in CLIP methods can be alleviated with the guidance of a pre-trained text encoder. In most datasets and few-shot settings, iCAR surpasses supervised-only, visual-linguistic only and CLIP baselines. We also notice a performance gap on FGVC Aircraft~\cite{aircraft} dataset. The reason may be that the text encoder do not have a strong ability to distinguish the names of these aircraft variants due to lacking of related data during pre-training, which could also be observed in the zero-shot classification experiments.

\end{document}